\documentclass[12pt,a4paper]{article}

\usepackage{graphicx}
\usepackage{amssymb}
\usepackage{amsmath}
\usepackage{amsfonts}
\usepackage{listings}
\usepackage{algorithm}
\usepackage{algpseudocode}
\usepackage{enumitem}
\usepackage{verbatim}
\usepackage{amsthm}
\usepackage{array}
\usepackage[utf8]{inputenc}
\usepackage{caption}
\usepackage{subcaption}
\usepackage{color}
\usepackage{url}

\newcommand{\hf}{{\textstyle{{\frac{1}{2}}}}}

\newcommand{\hfsig}{{\textstyle{{\frac{1}{2 \sigma^2}}}}}

\newcommand{\bx}{\mathbf{x}}
\newcommand{\bv}{\mathbf{v}}
\newcommand{\bz}{\mathbf{z}}
\newcommand{\by}{\mathbf{y}}
\newcommand{\beps}{\boldsymbol{\epsilon}}

\newcommand{\bepstheta}{\boldsymbol{\epsilon}_\theta}

\newcommand{\bmu}{\boldsymbol{\mu}}

\newcommand{\bzero}{\mathbf{0}}

\newcommand{\oa}{\overline{\alpha}}

\newcommand{\RR}{\mathbb{R}}

\newcommand{\bI}{\mathbf{I}}

\newcommand{\cN}{\mathcal{N}}

\begin{document}

\author{
Catherine F. Higham%
  \thanks{%
           School of Computing Science,
           University of Glasgow,
           Sir Alwyn Williams Building, Glasgow, G12 8QQ.
            Supported by EPSRC grant EP/T00097X/1. 
           (\texttt{Catherine.Higham@glasgow.ac.uk})
           }
           \and
        Desmond J. Higham%
            \thanks{%
           School of Mathematics,
           University of Edinburgh,
           EH9 3FD, UK.
            Supported by EPSRC grant EP/V046527/1. 
           (\texttt{d.j.higham@ed.ac.uk})
           }
 \and
        Peter Grindrod {\small CBE}%
        \thanks{%
        Mathematical Institute, University of Oxford, Oxford, United Kingdom. Supported  by EPSRC grant EP/R018472/1.
        (\texttt{grindrod@maths.ox.c.uk})
           }   
        }

\date{}
\title{Diffusion Models for Generative Artificial Intelligence: An Introduction for Applied Mathematicians
}

\maketitle
 \begin{abstract}
 
Generative artificial intelligence (AI) refers to algorithms that create synthetic but realistic output.
Diffusion models currently offer state of the art performance in generative AI for images.
They also form a key component in more general tools, including text-to-image generators and large language models.
Diffusion models work by adding 
noise to the available training data and then learning how to reverse the process.
The reverse operation may then be applied to new random data in order 
 to produce new outputs. 
 We provide a brief introduction to diffusion models for applied mathematicians and statisticians.
 Our key aims are (a) to present illustrative computational examples, (b) to 
 give a careful derivation of the underlying mathematical formulas involved, and (c) 
 to draw a connection with partial differential equation (PDE) diffusion models.
 We provide code for the computational experiments.
 We hope that this topic will be of interest to advanced undergraduate students and postgraduate students.
 Portions of the material may also provide useful motivational examples for those who teach  
 courses in stochastic processes, inference, machine learning, PDEs or scientific computing.
\end{abstract}

\section{Motivation}\label{sec:mot}

Generative artificial intelligence (AI) models are designed to create new outputs that are similar to the examples on which they were trained.
Over the past decade or so, advancements in generative AI have included the 
development of variational autoencoders \cite{Gir21,Kingma2014}, 
generative adversarial networks \cite{Ig23}
and 
transformers \cite{Tay22}.
In this work we focus on
denoising diffusion probabilistic models
 \cite{Ho20}; for simplicity we use the term diffusion models.
 They currently represent the state of the art in image generation \cite{Dhar21}, and form a key part of 
more sophisticated tools such as
DALL-E 2 and 3 \cite{ramesh2022hierarchical}.
We refer to 
\cite{Cao23,Cro23,Feu23,Po23},
and the references therein,
for details of the historical developments that have led to the current state-of-the art in generative AI.

The somewhat counterintuitive, but deceptively powerful, idea behind 
diffusion models is to destroy the training data by adding noise.
While doing so, the model learns how to reverse the process. In this way, the final model is able to start with new,
easily generated, random samples and denoise them, thereby 
generating new, synthetic examples.
The task of building and applying  
a simple, yet impressive, model can be described very succinctly---see Algorithms~\ref{alg:forward} and \ref{alg:backward} in section~\ref{sec:algs}.
However, deriving the expressions that go into these algorithms is not so straightforward, and we believe that there is a niche for a
careful and accessible mathematically-oriented treatment.

Our intended readership is advanced undergraduate students and postgraduate students in mathematics or relate disciplines. 
The material should be suitable for independent study, and there are many directions in which this material can be followed up---the literature
is rapidly expanding, and new extensions and connections are being discovered at a pace.
We also hope that portions of this material will provide higher education professionals with topical and engaging examples that can be slipped into 
courses on stochastics, numerics, PDEs or data science.

We aimed to keep the prerequisites to a minimum; these are
\begin{itemize}
\item for sections~\ref{sec:illust}--\ref{sec:algs} ideas from statistics: mean, variance, Gaussian distribution, Markov chains, conditional probability, 
\item for sections~\ref{sec:backwards} and \ref{sec:algs} ideas from deep learning: the stochastic gradient method, artificial neural networks,
\item for section~\ref{sec:pde} ideas from PDEs: multivariate calculus, the divergence theorem, spectral analysis.
\end{itemize}

We focus here on the task of image generation.
We describe a bare bones form of diffusion model,
explain carefully how the key mathematical expressions arise, and illustrate the concept via computational examples.
The key reference for this article is 
\cite{Ho20}, which built on
\cite{Sohl15} and 
received more than ten citations per day during the year 2023.
We also found \cite{Luo22}
to be a very useful resource.
  
In section~\ref{sec:illust} we present some pictures that 
give a feel for the idea of diffusion models in generative AI.
We then provide details of the relevant forward and backward processes 
in sections~\ref{sec:forwards} and \ref{sec:backwards}, respectively, which leads to the 
algorithms presented in section~\ref{sec:algs}.

We emphasize that this is a very  
active and fast-moving research topic with connections to many related areas.
In section~\ref{sec:further} we provide some links to the relevant literature.
That section also highlights wider issues around performance evaluation, computational expense, copyright, privacy, ethics, bias,
explainability and robustness.

We finish in section~\ref{sec:pde} with more speculative material that suggests a connection between 
stable diffusion models and deterministic PDEs, providing a link to more traditional applied mathematics.

\section{Illustration}\label{sec:illust}

A diffusion model \cite{Ho20} aims to generate realistic-looking images.
It works by 
\begin{description}
\item[(i)] taking an existing image and iteratively adding noise until the original information is lost, 
\item[(ii)] learning how to reconstruct the original image by iteratively removing the noise. 
\end{description}
After training, we can then use the reverse diffusion process to generate a realistic image from a new, random, starting point---remove the noise and see what emerges.

One way to conceptualize this method is to imagine an (unknown) probability distribution over the collection of all natural images.
We hope to sample from this distribution; more likely images should be chosen with higher probability.
We don't have access to this magic probability distribution. Instead, we have training data; that is, examples of natural images. 
We also have a pseudorandom number generator that allows us to sample from a standard Gaussian distribution.
In item (i) above, we are doing the easy part, mapping from the image distribution to the Gaussian distribution.
In item (ii) we learn the inverse operator, mapping from the Gaussian distribution to the image distribution.
This allows us to convert Gaussian samples into images.

\begin{figure}
    \centering
    \includegraphics[width=0.7\textwidth]{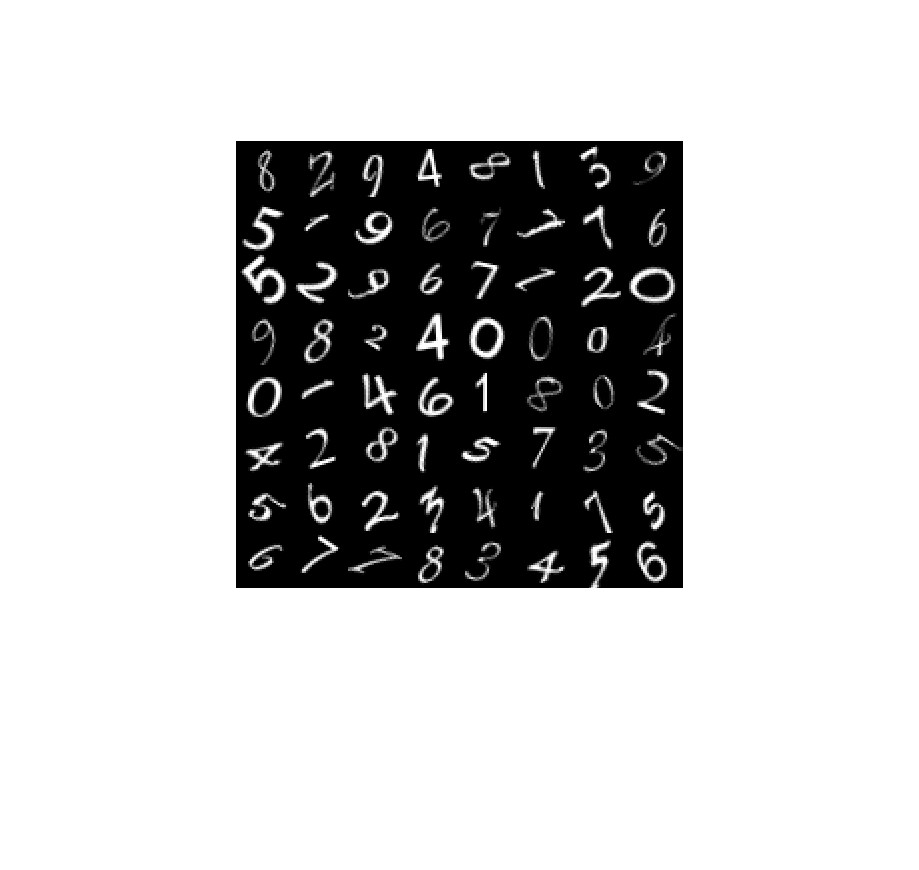}
    \caption{Representative set of $64$ images from the MNIST data set \cite{lcb-digits_old}.
    }
    \label{fig:mnist}
\end{figure}

We illustrate the idea of a diffusion model trained on images from the widely studied MNIST data set \cite{lcb-digits_old}.
Here, each image represents a handwritten digit from the set  $\{0,1,2,\ldots,9\}$.
These low resolution images are black-and-white with $28 \times 28$ pixels, resized by the model to $32 \times 32$.
 Figure~\ref{fig:mnist} shows a representative collection of $64$ images.

\begin{figure}
    \centering
    \includegraphics[width=0.7\textwidth]{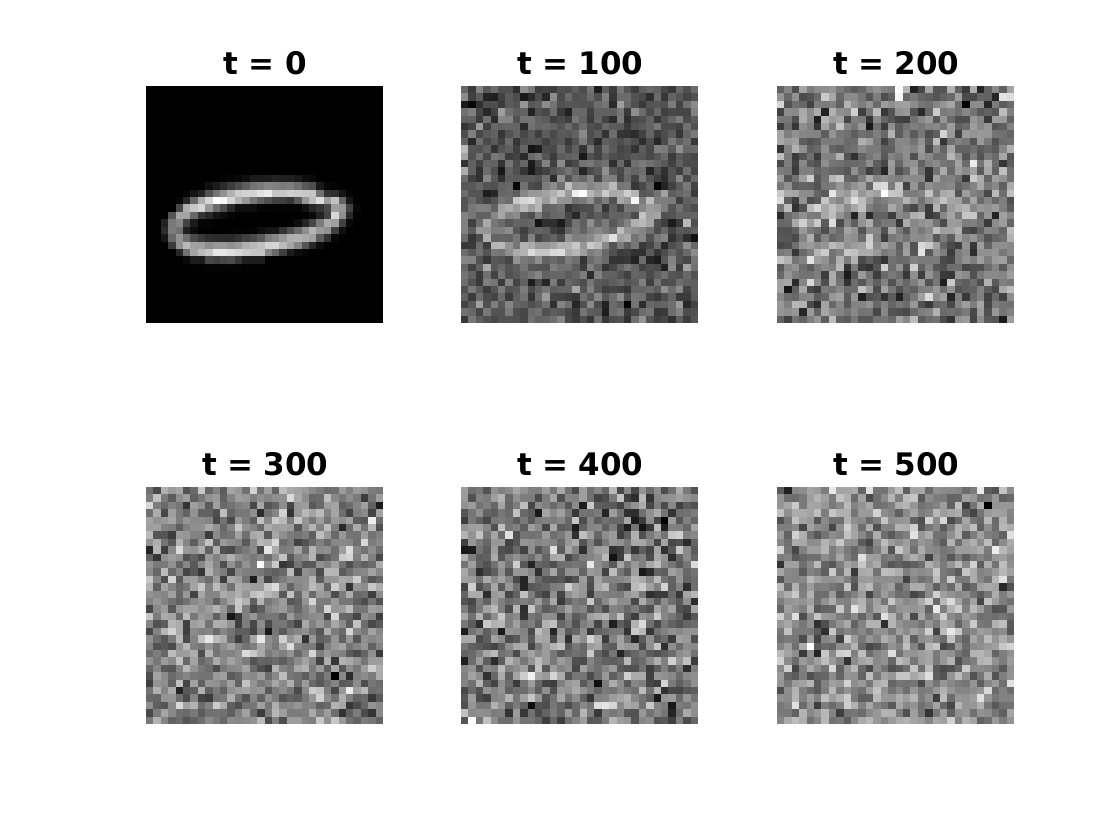}
    \caption{Result of forward map noising over time.
    }
    \label{fig:forward}
\end{figure}

\begin{figure}
    \centering
    \includegraphics[width=0.35\textwidth]{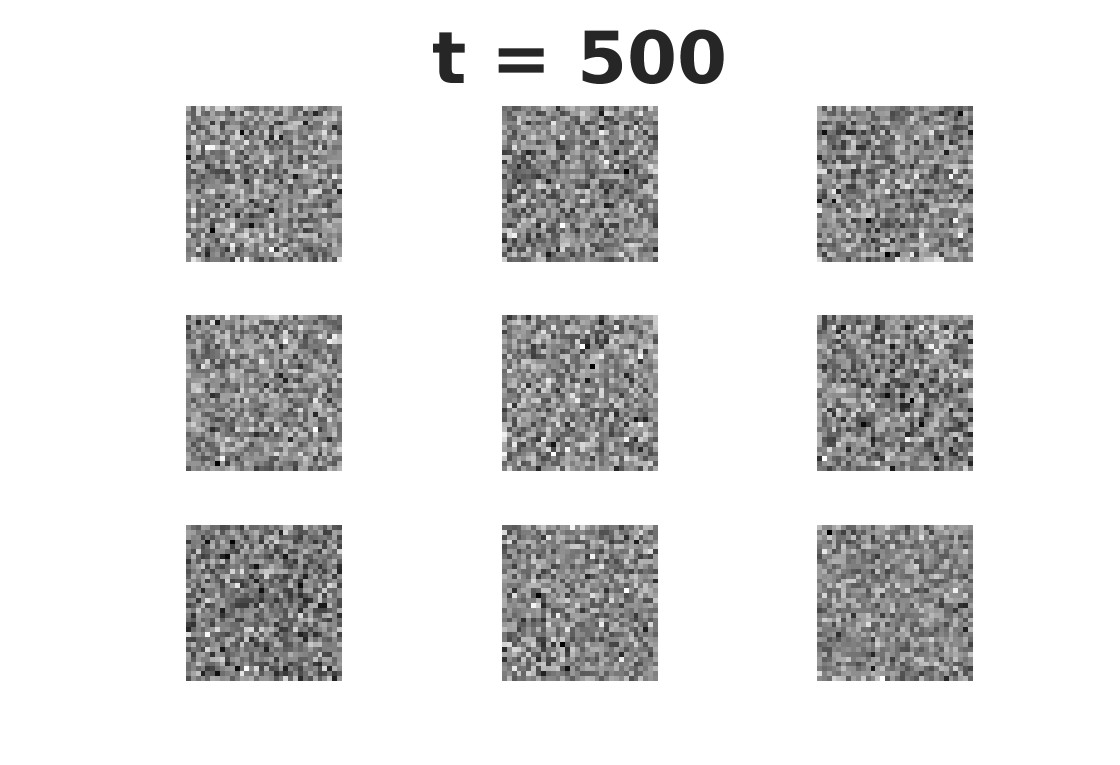}
    \includegraphics[width=0.35\textwidth]{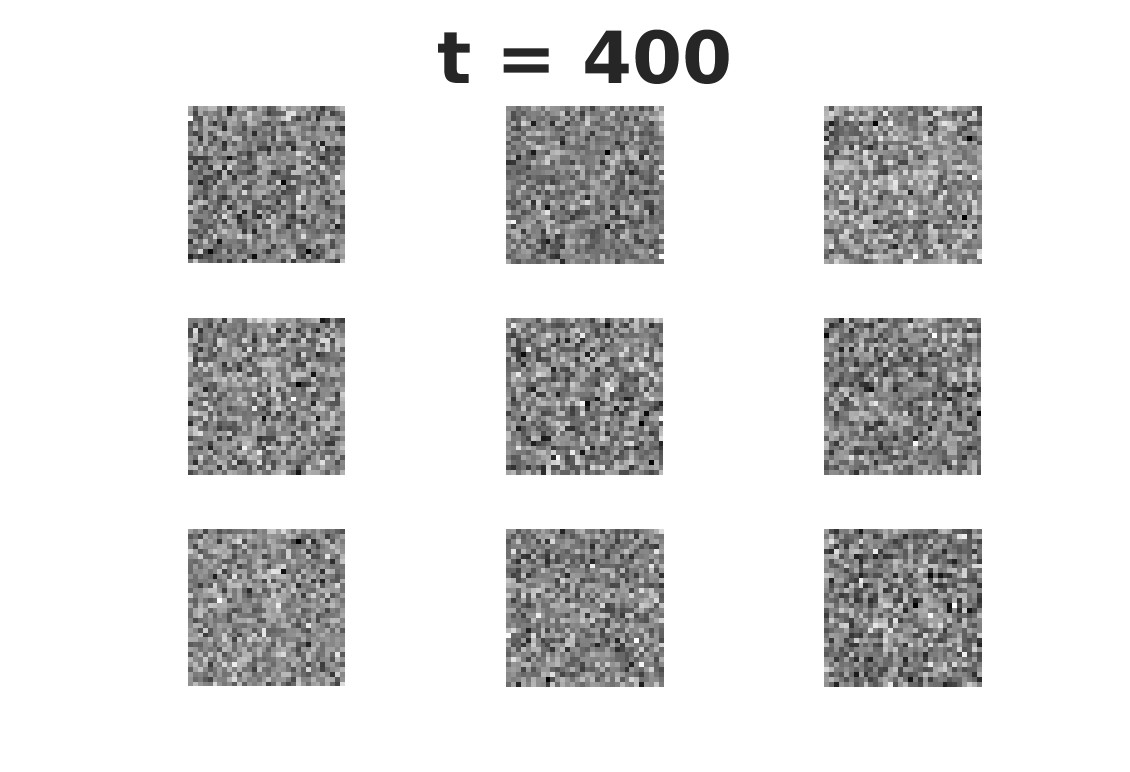}\\
    \includegraphics[width=0.35\textwidth]{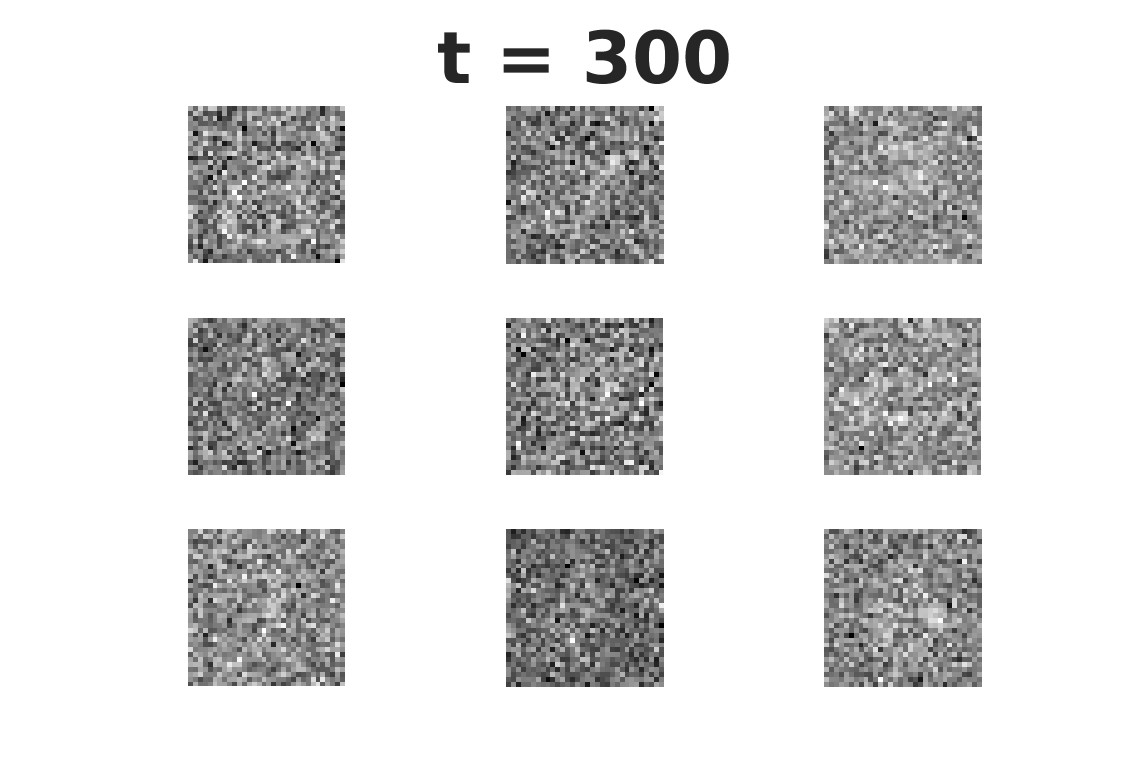} 
    \includegraphics[width=0.35\textwidth]{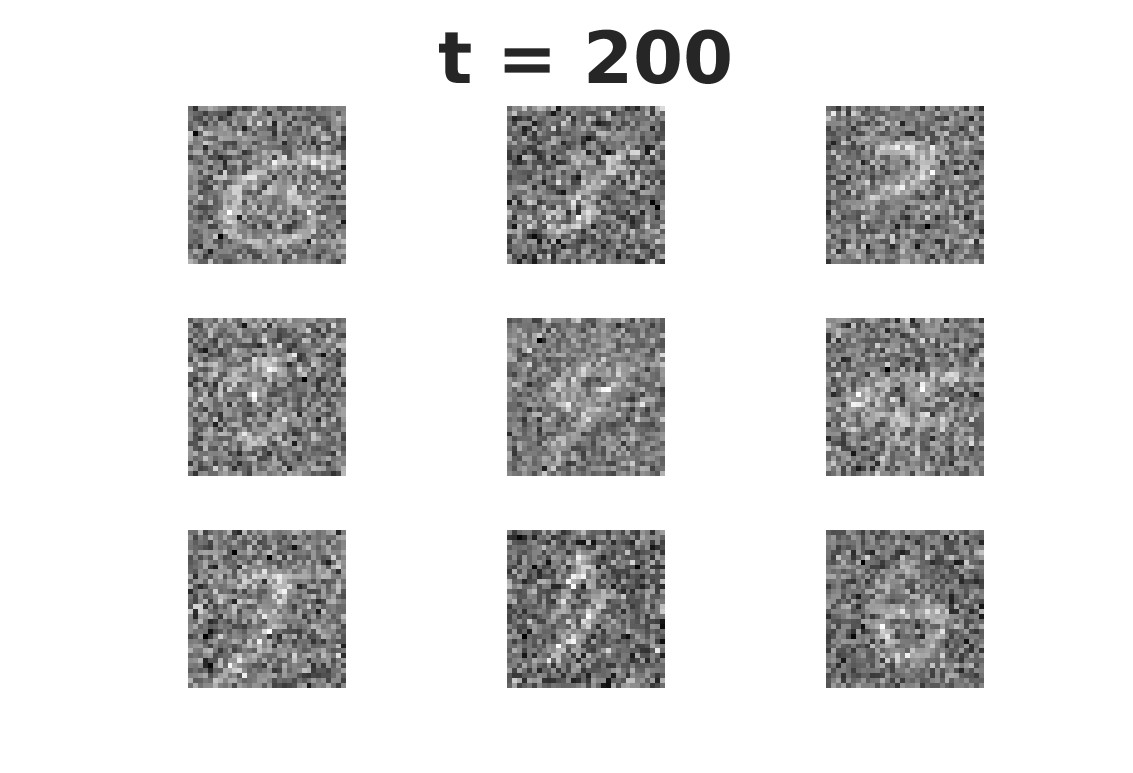}\\
    \includegraphics[width=0.35\textwidth]{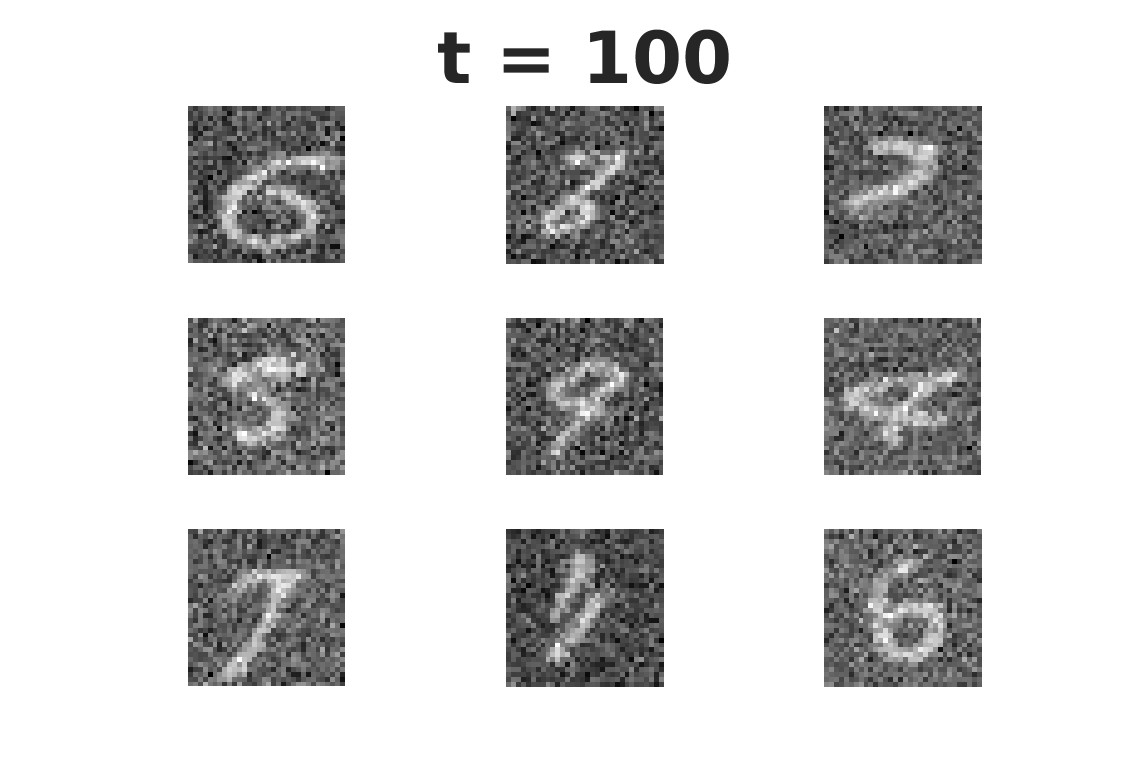}
    \includegraphics[width=0.35\textwidth]{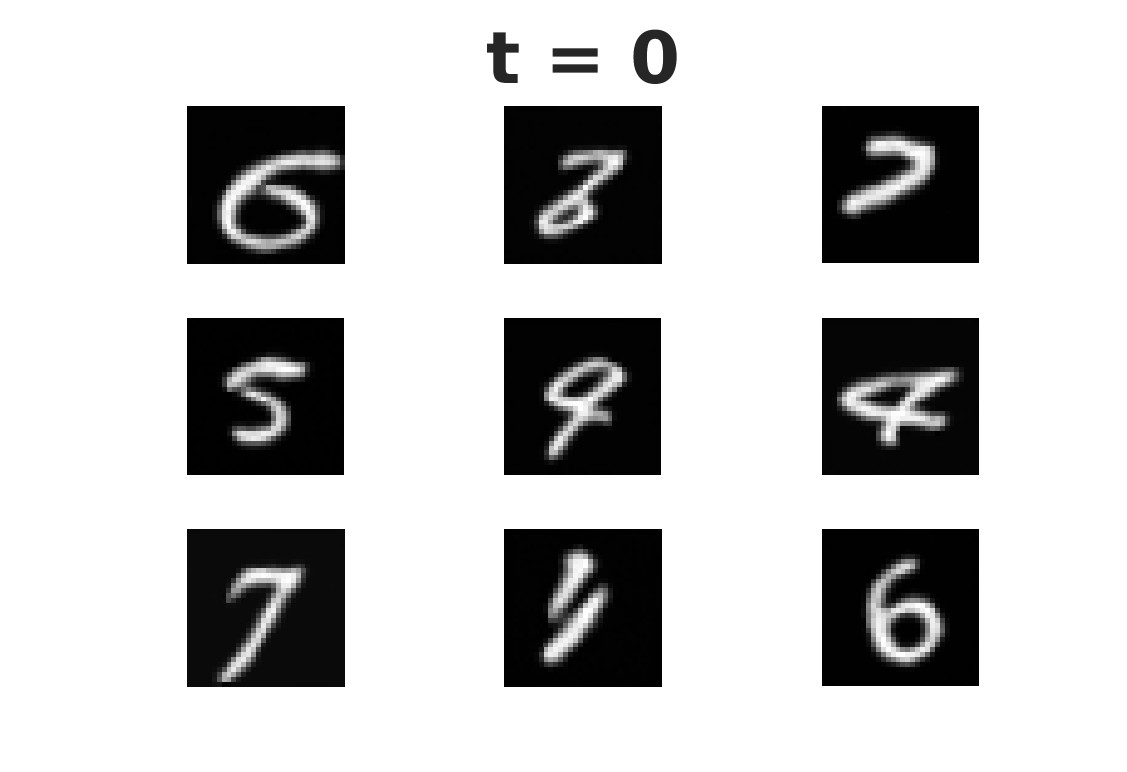}\\
    \caption{Result of backward map denoising over time, for $9$ different random choices at $t=500$.
    }
    \label{fig:reverse}
\end{figure}

\begin{figure}
    \centering
    \includegraphics[width=0.35\textwidth]{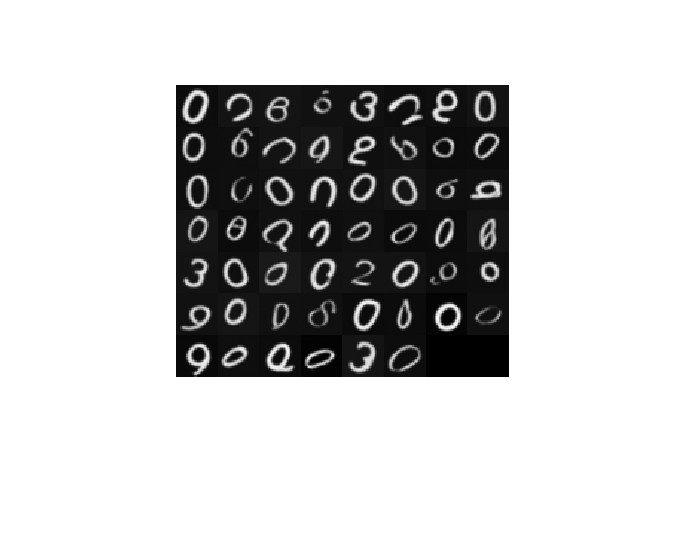}
    \includegraphics[width=0.35\textwidth]{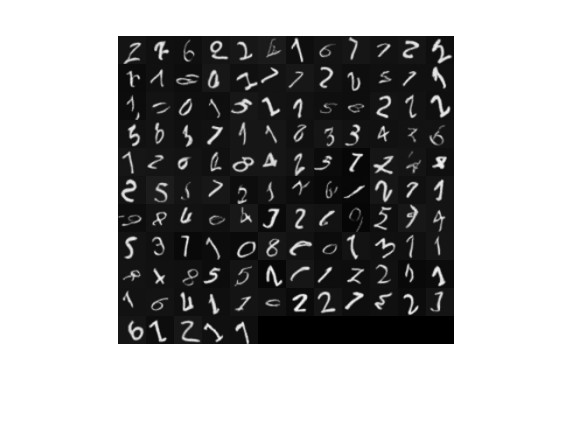}\\
    \includegraphics[width=0.35\textwidth]{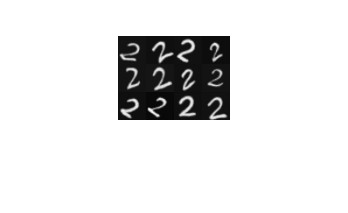} 
    \includegraphics[width=0.35\textwidth]{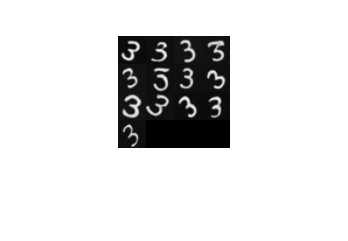}\\
    \includegraphics[width=0.35\textwidth]{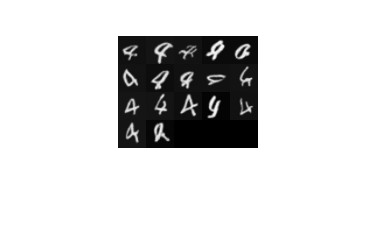}
    \includegraphics[width=0.35\textwidth]{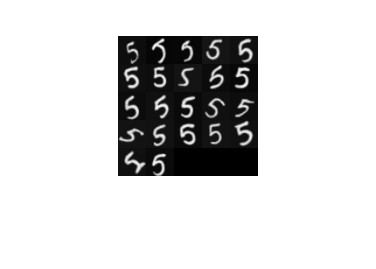}\\
    \includegraphics[width=0.35\textwidth]{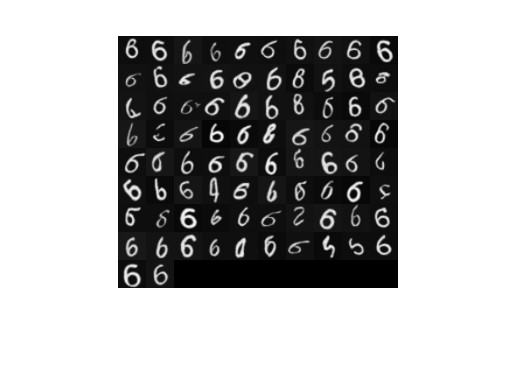}
    \includegraphics[width=0.35\textwidth]{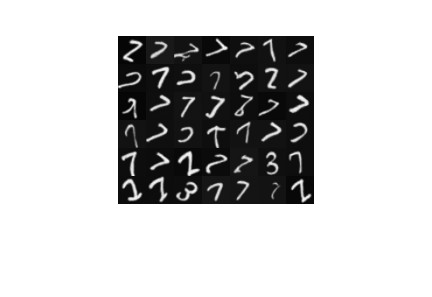}\\
    \includegraphics[width=0.35\textwidth]{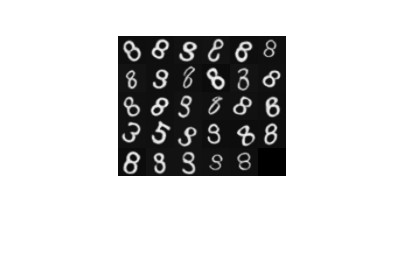}
    \includegraphics[width=0.35\textwidth]{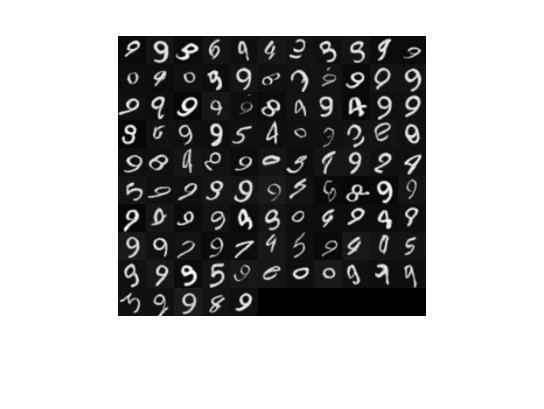}
    \caption{$500$ synthetic images generated by the diffusion model from different time $t=500$ noise samples. These have been sorted into classes by a 
    convolutional neural network classifier that was trained separately on real MNIST data.
    }
    \label{fig:clusters}
\end{figure}

 Figures~\ref{fig:forward}--\ref{fig:clusters} were produced with a diffusion model based on a Mathworks tutorial at 
 {\scriptsize
\begin{verbatim}
https://uk.mathworks.com/help/deeplearning/ug/generate-images-using-diffusion.html
\end{verbatim}
}
\noindent
Figure~\ref{fig:forward} illustrates the forward process that is used in the training phase.
At time $t=0$ we have an MNIST image.
At each integer time 
$t = 0,1,2,\ldots,499$ Gaussian noise is added. At $t=500$ there is no visible 
evidence of the original image.

Figure~\ref{fig:reverse} shows the effect of the backward process that is available after training.
The top left panel displays nine randomly chosen final time $t=500$ images---pure noise matrices consisting of independent Gaussian samples.
We show the effect of applying the backward, denoising process as time is reversed.
At $t=0$ the model has produced new, synthetic examples that, in at least eight of the nine cases, correspond to handwritten digits.
We emphasize that labels were not used in the training process. In this simple, unconditional model there is no way to control which (if any) of the $t=0$ images will resemble any particular category of digit.

In Figure~\ref{fig:clusters} we show the results from a larger experiment. Here we used the trained diffusion model to generate images from 
500 independent time $t=500$ choices.
For this figure, we separated the images into categories using an independent convolutional neural network classifier that was trained 
separately on real MNIST data. 
Since we have no control over how many of the 500 images will appear in each class, the number of synthetic outputs in each category varies 
considerably.

\begin{figure}
    \centering
    \includegraphics[width=0.35\textwidth]{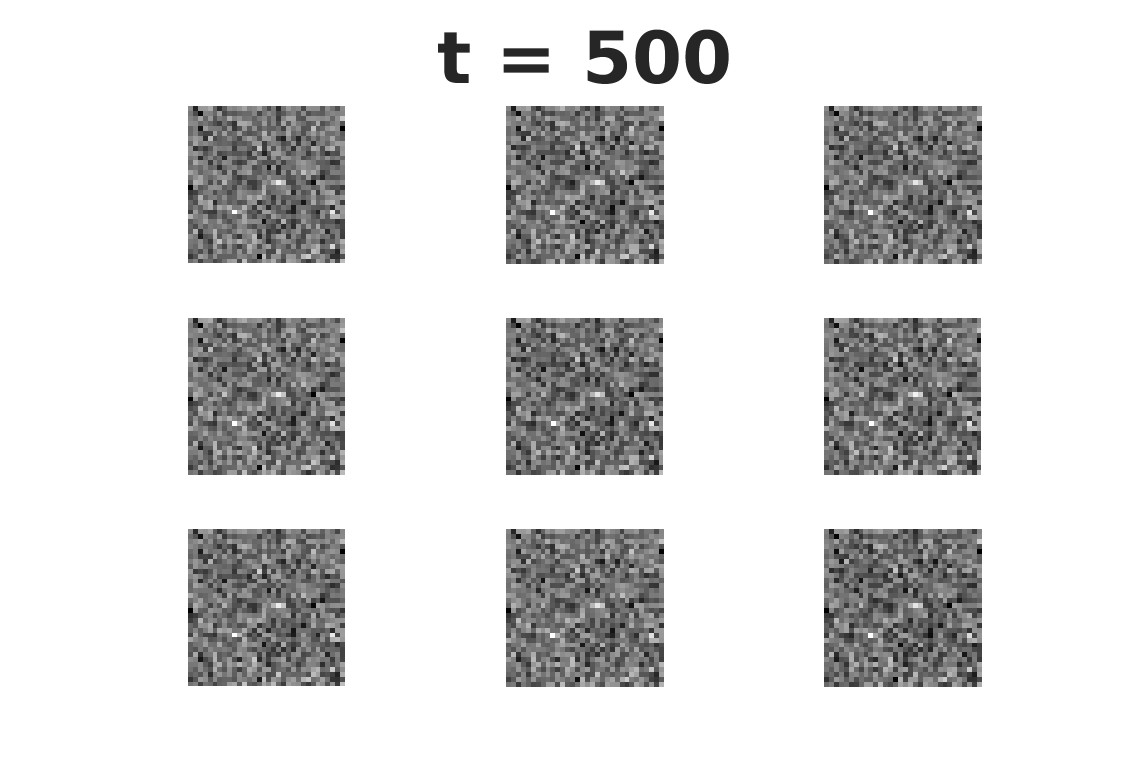}
    \includegraphics[width=0.35\textwidth]{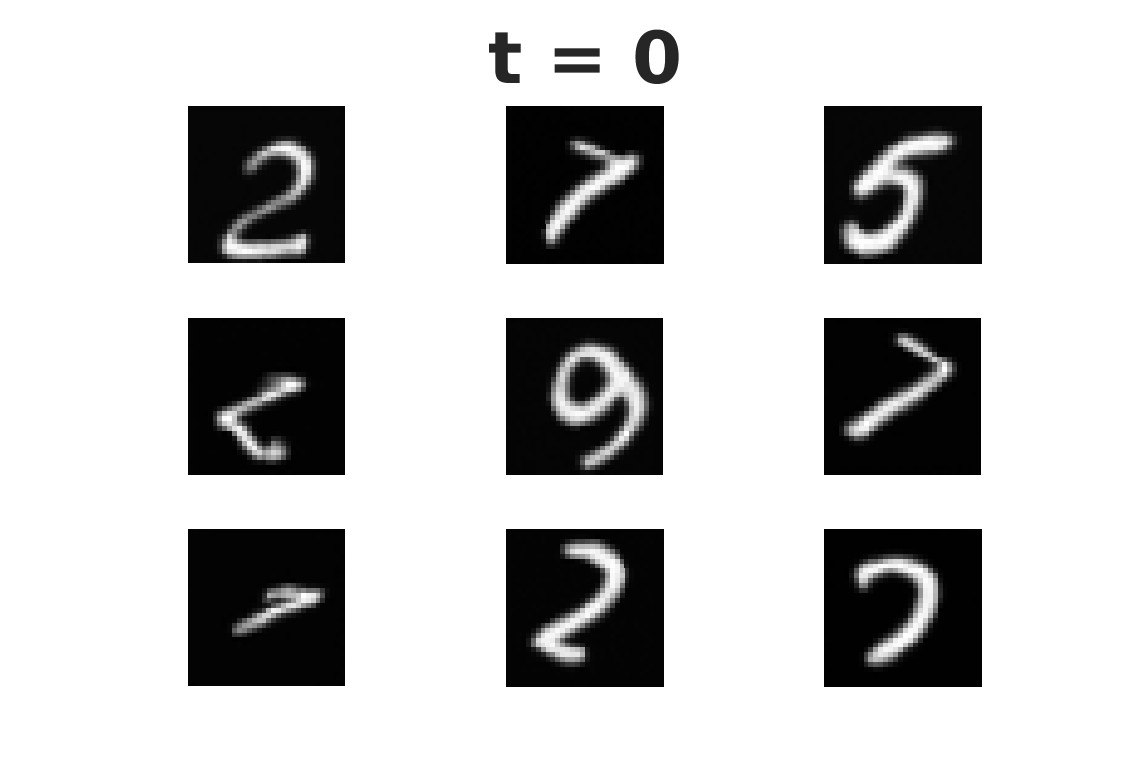}\\
     \caption{Nine images (right) created from the same Gaussian noise at $t=500$ (left).
    }
    \label{fig:sameGN}
\end{figure}

\begin{figure}
    \centering
    \includegraphics[width=0.35\textwidth]{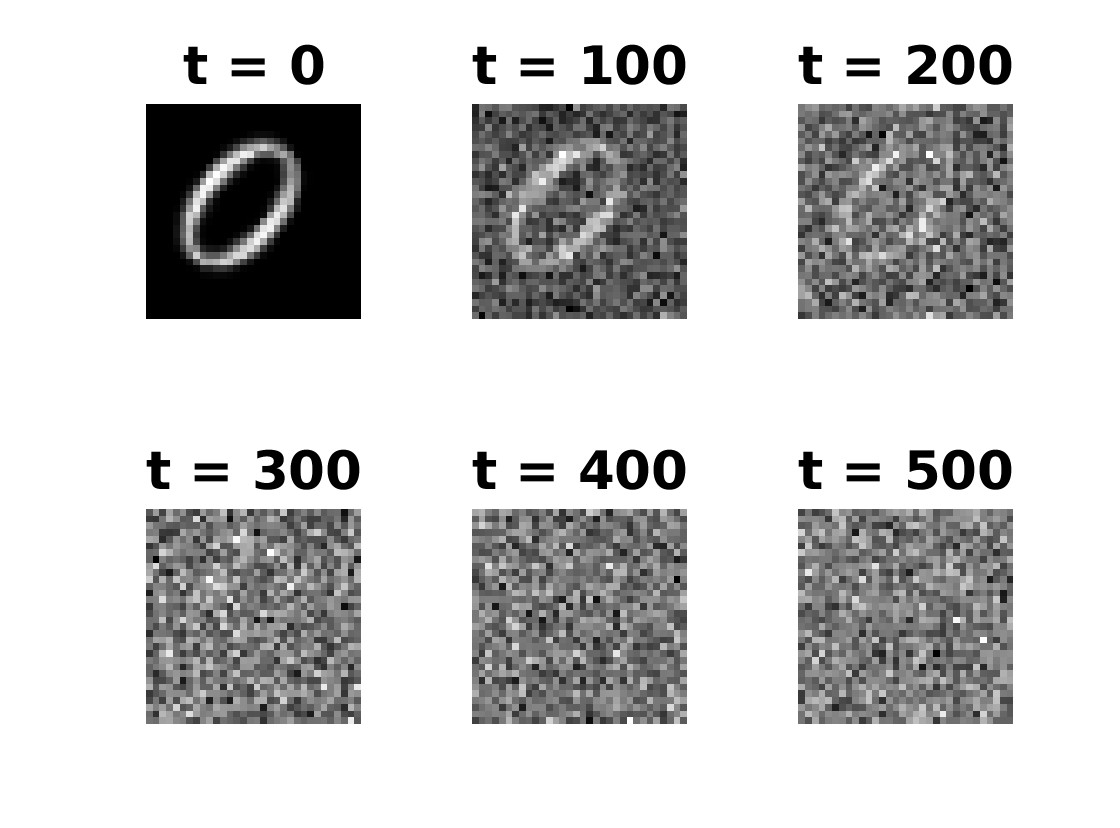}
    \includegraphics[width=0.35\textwidth]{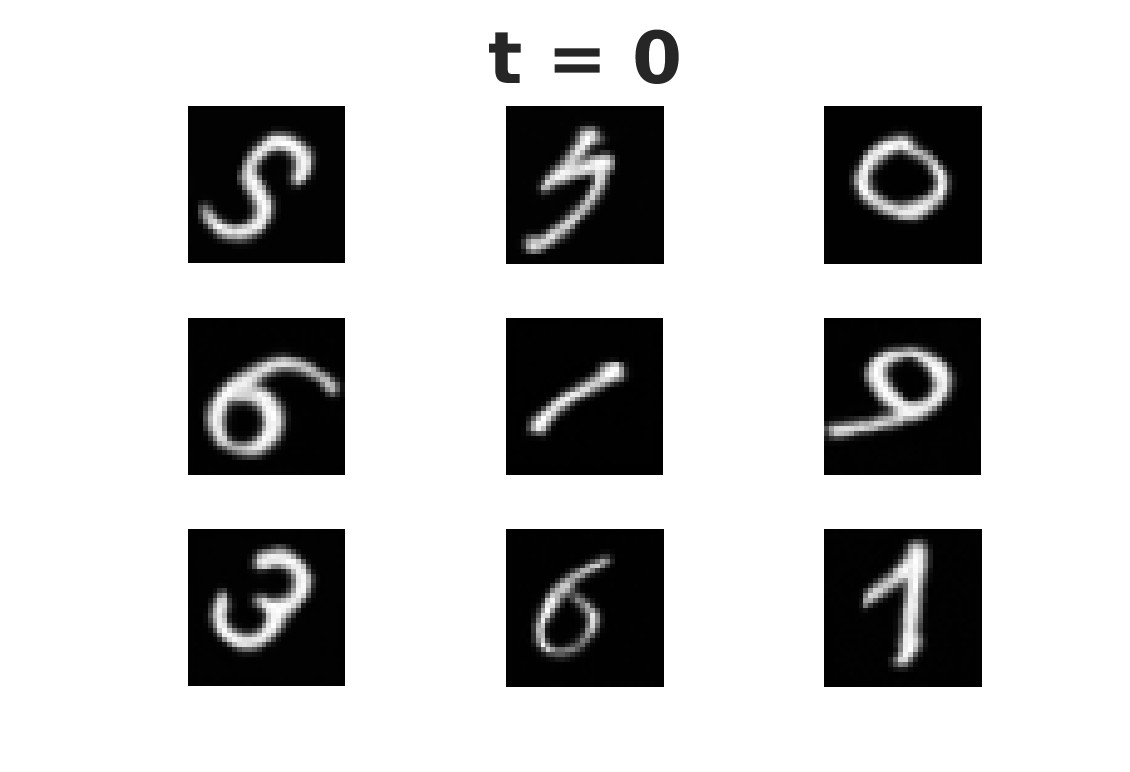}\\
     \caption{Nine images (right) created from the noisy version (left $t=500$) of one original image (left $t=0$).
    }
    \label{fig:addGN}
\end{figure}

We finish with two experiments which illustrate that the backward, denoising process is both stochastic and unpredictable.
In Figure~\ref{fig:sameGN} we show the images generated after applying nine independent denoising runs to the same 
Gaussian at $t=500$. We see that the denoising process can produce considerably different results from a single source of 
randomness. In Figure~\ref{fig:addGN} we perform a similar experiment where the $t=500$ data emerges from the training set.
On the left we show a training image undergoing the forward, noising process up to time $t=500$.
On the right we show the results from nine independent denoising runs on this time $t=500$ data.
We see that none of the synthetically generated images resemble the original.

\section{Forwards}\label{sec:forwards}

We begin this section with some background on Gaussian random variables; see a standard text such as 
\cite{bish007,Murph22} for more details.
When dealing with Gaussians, we will always consider the  multivariate, isotropic case.
We denote the probability density at a point $\bx \in \RR^{d}$ by 
$\cN( \bx; \bmu, \sigma \bI)$,
where 
\begin{equation}
\cN( \bx; \bmu, \sigma \bI) := 
  \frac{1}{ \sigma (2 \pi)^{1/d} }
  \exp\left( -\hfsig (\bx - \bmu)^T (\bx - \bmu)  \right).
  \label{eq:CN}
\end{equation}
Here, $\bmu \in \RR^{d}$ is the mean, and we will refer to $\sigma^2$ as the variance, since the corresponding 
covariance matrix has the form 
$\sigma^2 \bI$, with $\bI \in \RR^{d \times d}$ denoting the identity matrix.
Such Gaussian random variables have the important property that their sums remain Gaussian, with means and variances
combining additively: the sum of two independent 
Gaussians with means $\bmu_1$ and $\bmu_2$ and variances 
$\sigma_1^2$ and 
$\sigma_2^2$ is a Gaussian random variable with mean 
$\bmu_1 + \bmu_2$ and variance 
$\sigma_1^2 + \sigma_2^2$.
The term standard Gaussian refers to the case where the mean is $\bzero \in \RR^d$ and the variance is $1$.
Multiplying a standard Gaussian by the scalar $\sigma$
and shifting by $\bmu \in \RR^{d}$ 
produces a Gaussian with mean $\bmu$ and variance $\sigma^2$.
It follows that if $\by$ and $\bz$ are independent standard Gaussians, and $a$ and $b$ are scalars, then 
$a\by + b\bz$ is Gaussian with mean zero and variance $a^2 + b^2$; so 
$a\by + b\bz$ can be sampled as
$\sqrt{a^2+b^2} \, \beps$, where $\beps$ is a standard Gaussian.

We consider images that can be described by $d$ real numbers, typically pixel values, and we collect these into a vector in  
$\RR^d$. In practice pixel values might be constrained---for example only integers between $0$ and $255$ might be allowed---but
we ignore this issue here for simplicity.

Given an image $\bx_{0} \in \RR^d$, the \emph{forward process} iteratively adds noise to create a sequence
$\bx_{1}, \bx_{2}, \ldots, \bx_{T}$ according to the rule 
\begin{equation}
     \bx_{t} = \sqrt{1 - \beta_t} \, \bx_{t-1} + \sqrt{ \beta_t } \, \beps_t.
     \label{eq:f1}
\end{equation}
Here, each $\beps_t$ is an independent standard Gaussian and the scalar parameter $\beta_t$ is between zero and one.
The sequence $\beta_1, \beta_2, \ldots, \beta_{T}$, known as the \emph{variance schedule}, is predetermined.
For example, in \cite{Ho20}, linearly increasing values from $\beta_1 = {10}^{-4}$ to $\beta_T = 0.02$ are used.
Since $\beta_t$ here is increasing, more noise is added as the forward process evolves.
It is useful to think of $t$ as a time-like variable. At time zero we have an image and at time $T$ we effectively have 
pure Gaussian noise. 

The process (\ref{eq:f1}) defines a discrete time Markov process, and the associated transition density may be written
\begin{equation}
q( \bx_t \, | \, \bx_{t-1} ) = 
    \cN( \bx_t; \sqrt{ 1- \beta_t} \, \bx_{t-1}, \beta_t \bI).
\label{eq:ftrans}
\end{equation}
This quantifies the probability of observing $\bx_t$ at time $t$, given $\bx_{t-1}$ at time $t-1$. 

Updating over one time step in  the forward process (\ref{eq:f1}) is straightforward; just scale the current value and add Gaussian noise.
For later use, it is helpful to know that stepping from time zero to a general time $t$ is possible with a single leap.
To see this, we introduce
$
\alpha_t = 1 - \beta_t
$
so that 
\begin{equation}
     \bx_{t} = \sqrt{\alpha_t} \, \bx_{t-1} + \sqrt{ 1 - \alpha_t } \, \beps_t.
     \label{eq:bxalpha}
     \end{equation}
Then, applying (\ref{eq:bxalpha}) again, we have 
\begin{eqnarray}
       \bx_{t} &=& 
           \sqrt{\alpha_t} \left( 
            \sqrt{\alpha_{t-1}} \, \bx_{t-2} + \sqrt{ 1 - \alpha_{t-1} } \, \beps_{t-1}
          \right)
        + \sqrt{ 1 - \alpha_t } \, \beps_t \nonumber \\
        &=& 
            \sqrt{\alpha_t \alpha_{t-1}} \, \bx_{t-2} +
              \sqrt{\alpha_t} \sqrt{ 1 - \alpha_{t-1} } \, \beps_{t-1}
        + \sqrt{ 1 - \alpha_t } \, \beps_t.
        \label{eq:bx2}
\end{eqnarray}
Using the properties of Gaussians mentioned at the start of this section, we see that $\sqrt{\alpha_t} \sqrt{ 1 - \alpha_{t-1} } \, \beps_{t-1}
        + \sqrt{ 1 - \alpha_t } \, \beps_t   $
        can be combined into a single Gaussian. In this way, 
        (\ref{eq:bx2}) may be written 
  \[
  \bx_{t}
   = 
    \sqrt{\alpha_t \alpha_{t-1}} \, \bx_{t-2} + 
     \sqrt{1 - \alpha_t  \alpha_{t-1}} \, \beps_{t,t-2},
  \]      
  where 
  $\beps_{t,t-2}$ is a standard Gaussian.

  Proceeding inductively, suppose
  that for some $k$ between $t-2$ and $1$
  \begin{equation}
  \bx_{t}
   = 
    \sqrt{\alpha_t \alpha_{t-1} \ldots \alpha_{k+1}} \, \bx_{k} + 
     \sqrt{1 - \alpha_t  \alpha_{t-1}\ldots \alpha_{k+1}  } \, \epsilon_{t,k},
     \label{eq:bxind}
  \end{equation}
where $ \epsilon_{t,k}$ is a standard Gaussian. Then replacing $\bx_{k}$ using (\ref{eq:bxalpha}) 
\[
  \bx_{t}
   = 
    \sqrt{\alpha_t \alpha_{t-1} \ldots \alpha_{k+1}} \, 
     \left( 
    \sqrt{\alpha_{k} } \, \bx_{k-1} + \sqrt{ 1 - \alpha_{k} } \, \beps_{k}
    \right)
     + 
     \sqrt{1 - \alpha_t  \alpha_{t-1}\ldots \alpha_{k+1}  } \,  \epsilon_{t,k}.
  \]
  Again replacing the sum of two independent Gaussians by a single, appropriate Gaussian, we have
  \begin{eqnarray*}
  \bx_{t}
   &= &
    \sqrt{\alpha_t \alpha_{t-1} \ldots \alpha_{k} }\, 
      \bx_{k-1} + 
      \sqrt{   \alpha_t \alpha_{t-1} \ldots  \alpha_{k+1}(1-\alpha_{k}) + 1 - \alpha_t  \alpha_{t-1}\ldots \alpha_{k+1}}  \, 
      \epsilon_{t,k-1},\\
      &=&
       \sqrt{\alpha_t \alpha_{t-1} \ldots \alpha_{k} } \, 
      \bx_{k-1} + 
      \sqrt{ 1 -  \alpha_t \alpha_{t-1} \ldots  \alpha_{k}  }  \, 
      \epsilon_{t,k-1},
      \end{eqnarray*}
      where $\epsilon_{t,k-1}$ is a standard Gaussian. 
      Hence, the form (\ref{eq:bxind}) is valid all the way down to $k=0$. So,
      letting 
      \begin{equation}
        \oa_t = \prod_{i=1}^{t} \alpha_i,
         \label{eq:ahatdef}
       \end{equation}
we may write
\begin{equation}
  \bx_{t}
   = 
    \sqrt{\oa_t} \, \bx_{0} + 
     \sqrt{1 - \oa_{t}} \, \bar{\beps}_t,
   \label{eq:f2}
\end{equation}   
  where 
  $\bar{\beps}_t$ is a standard Gaussian.
  We may therefore step directly from time $0$ to any later time $t$ using a single Gaussian.
  This proves convenient for the analysis in section~\ref{sec:backwards} and also for the 
  training algorithm discussed in section~\ref{sec:algs}.

In terms of a transition density, (\ref{eq:f2}) shows that 
\begin{equation}
 q(\bx_{t} \, | \, \bx_{0} ) := \cN( \bx_{t}; \sqrt{ \oa_t } \, \bx_{0}, (1 - \oa_t) \bI).
   \label{eq:xtzerotrans}
\end{equation}

\section{Backwards}\label{sec:backwards}
We now consider the reverse process.
We are interested in the probability of $\bx_{t-1}$ given $\bx_t$ and $\bx_{0}$; that is,
$q( \bx_{t-1} \, | \, \bx_{t}, \bx_{0} ) $. 
To proceed we will make use of a result in conditional probability theory known as the product rule,
\cite{bish007,Murph22}, which for our purposes may be written
\[
  P(A,B,C) = P(A \, | \, B,C) \, P(B,C) = P(A \, | \, B,C) \, P(B\,| \, C) \, P(C).
\]
By symmetry, we also have 
\[
P(A,B,C) =  P(B,A,C) =  P(B \, | \, A,C) \, P(A,C) = P(B \, | \, A,C) \, P(A\,| \, C) \, P(C).
\]
Hence,
\[
P(A \, | \, B,C)
=
\frac{
P(B \, | \, A,C)
\,
P(A\,| \, C)
}
{
P(B\,| \, C)
}.
\]
We will use this in the form 
\begin{eqnarray}
 q( \bx_{t-1} \, | \, \bx_{t}, \bx_{0} ) 
   &=& 
     \frac{  q( \bx_{t} \, | \, \bx_{t-1}, \bx_{0} )  \, q( \bx_{t-1} \, | \,  \bx_{0} ) } {  q( \bx_{t} \, | \,\bx_{0} )}.
     \label{eq:fromBayes}
     \end{eqnarray}
So now we focus on the quantities appearing on the right hand side of (\ref{eq:fromBayes}).

By the Markovian nature of the forward process, from (\ref{eq:ftrans}),
\begin{equation}
q( \bx_t \, | \, \bx_{t-1}, \bx_{0} ) = 
q( \bx_t \, | \, \bx_{t-1} ) = 
    \cN( \bx_t; \sqrt{\alpha_t} \,\bx_{t-1}, (1-\alpha_t) \bI).
\label{eq:fA}
\end{equation}
Making use of (\ref{eq:xtzerotrans}) for $\bx_{t}$ and $\bx_{t-1}$,  we then see that 
\begin{equation}
 q( \bx_{t-1} \, | \, \bx_{t}, \bx_{0} ) 
 =
 \frac{
 \cN( \bx_t; \sqrt{\alpha_t} \,\bx_{t-1}, (1-\alpha_t) \bI) \, 
\cN( \bx_{t-1}; \sqrt{ \oa_{t-1}} \, \bx_{0}, (1 - \oa_{t-1}) \bI)
 }
 {
 \cN( \bx_{t}; \sqrt{ \oa_t } \, \bx_{0}, (1 - \oa_t) \bI)
 }.
    \label{eq:bigarat}
\end{equation}
From the definition (\ref{eq:CN}), and 
ignoring the normalizing constants, we see that this expression has the form 
\begin{eqnarray}
&&\exp \left(
-
\hf
\frac{
(\bx_t - \sqrt{\alpha_t} \, \bx_{t-1} )^T  (\bx_t - \sqrt{\alpha_t} \, \bx_{t-1} ) 
}
{
1 - \alpha_t
}
-
\hf
\frac{
(\bx_{t-1} - \sqrt{\oa_{t-1}} \, \bx_{0} )^T  (\bx_{t-1} - \sqrt{\oa_{t-1}} \, \bx_{0} ) 
}
{
1 - \oa_{t-1}
} 
\right.
\nonumber 
\\
&&
\left.
\mbox{}
+ 
\hf
\frac{
(\bx_{t} - \sqrt{\oa_{t}} \bx_{0} )^T  (\bx_{t} - \sqrt{\oa_t} \, \bx_{0} ) 
}
{
1 - \oa_t
}
\label{eq:Bayesrat}
\right).
\end{eqnarray}
We will show that this expression matches 
\begin{equation}
 \cN \left( \bx_{t-1}; \mu_q(\bx_{t},\bx_0), \sigma^2_q(t) \bI \right), 
\label{eq:Nmq}
\end{equation}
for appropriate $\mu_q(\bx_{t},\bx_0)$ and $\sigma^2_q(t)$.
From (\ref{eq:CN}), we can find $\sigma^2_q(t)$ by considering the coefficient of 
$\mbox{}-\bx_{t-1}^T \bx_{t-1}$ in the exponent of (\ref{eq:Bayesrat}). This coefficient is given by 
\[
\hf \frac{\alpha_t}{1 - \alpha_t}
+ 
\hf \frac{1}{1 - \oa_{t-1}}
=
\hf \frac{ \alpha_t ( 1 - \oa_{t-1}) + 1 - \alpha_t}{(1 - \alpha_t)( 1 - \oa_{t-1})}
=
\hf\left(
\frac{ 1 - \oa_t }  {(1 - \alpha_t)( 1 - \oa_{t-1}) } \right),
\]
where we used $\alpha_t \, \oa_{t-1} = \oa_{t}$ from (\ref{eq:ahatdef}).
Hence, 
\begin{equation}
\sigma^2_q(t) =  \frac{(1 - \alpha_t) ( 1 - \oa_{t-1}) } { 1 - \oa_t }.
\label{eq:sigmasq}
\end{equation}
Using the functional form (\ref{eq:CN}) again,
we can find $\mu_q(\bx_{t},\bx_0)$ by considering the vector, say $\bv$, such that 
$\bx_{t-1}^T \bv$ is the cross-product in the exponent of (\ref{eq:Bayesrat}). We see that 
\[
\frac{\mu_q(\bx_{t},\bx_0)  } {\sigma^2_q(t) }
=
\bv = 
 \frac{\sqrt{\alpha_t} \, \bx_{t}}{1 - \alpha_t}
 +\frac{\sqrt{\oa_{t-1}}\, \bx_{0}}{1 - \oa_{t-1}}.
\]
Hence, using (\ref{eq:sigmasq}),
\begin{equation}
\mu_q(\bx_{t},\bx_0)  
=
\frac{ \sqrt{\alpha_t} \, ( 1 - \oa_{t-1})\bx_{t} + \sqrt{ \oa_{t-1} }   (1 - \alpha_t) \bx_0}  {1-\oa_t}.
\label{eq:muqexp}
\end{equation}

We wish to compute a sample from the distribution in 
(\ref{eq:Nmq}).
This will allow us to perform the required transition along the backwards process.
Our approach is to estimate the mean in (\ref{eq:Nmq}) and then shift with an appropriate Gaussian in order to match the required variance.

If we know $\bx_{t}$ and $\bar{\beps}_t$ in (\ref{eq:f2}) then we may write 
\[
 \bx_{0} = \frac{  \bx_{t} - \sqrt{ 1 - \oa_{t} } \, \bar{\beps}_t}  { \sqrt{ \oa_{t} }}.
 \]
 Substituting this expression for $\bx_{0}$ into (\ref{eq:muqexp}) we see that the mean of $\bx_{t-1}$, given 
 $\bx_{t}$ and $\bx_{0}$, takes the form 
 \begin{equation}
 \mu_q(\bx_{t},\bx_0) = 
  \frac{ \sqrt{\alpha_t} ( 1 - \oa_{t-1}) } {1-\oa_t} \bx_{t}
  +
 \frac{ \sqrt{ \oa_{t-1}}  (1 - \alpha_t) } {(1-\oa_t) \sqrt{ \oa_{t} }}  \bx_{t}
 -
 \frac{ \sqrt{ \oa_{t-1}} (1 - \alpha_t) \sqrt{1-\oa_t}}{ (1-\oa_t)\sqrt{ \oa_{t} }} \,
 \bar{\beps}_t.
 \label{eq:muq2}
    \end{equation}
    Noting from (\ref{eq:ahatdef}) that 
    $\oa_{t-1}/\oa_{t} = 1/\alpha_t$ and 
     $\alpha_t \times \oa_{t-1} = \oa_{t}$, we find that 
    in (\ref{eq:muq2}) the coefficient of $\bx_{t}$ simplifies as follows: 
    \[
     \frac{ \sqrt{\alpha_t} ( 1 - \oa_{t-1}) } {1-\oa_t }
      +
      \frac{ \sqrt{ \oa_{t-1}}  (1 - \alpha_t) }  { (1-\oa_t)  \sqrt{ \oa_{t} }}
      =
      \frac{1}{ \sqrt{\alpha_t}   (1-\oa_t)  }
       \left(
          \alpha_t ( 1 - \oa_{t-1}) 
          + 1 - \alpha_t
         \right)
      = \frac{1}{\sqrt{\alpha_t}}.
      \]
      Similarly, the coefficient of $ \bar{\beps}_t $  in (\ref{eq:muq2}) simplifies to
      \[
      \mbox{} - \frac{ \sqrt{ \oa_{t-1}} (1 - \alpha_t) \sqrt{1-\oa_t}}{ (1-\oa_t)\sqrt{ \oa_{t} }}
      =
      \mbox{} - \frac{ 1 - \alpha_t} {  \sqrt{\alpha_t} \sqrt{1-\oa_t} }.
      \]

  Hence, (\ref{eq:muq2}) may be written
   \begin{equation}
 \mu_q(\bx_{t},\bx_0) = 
\frac{ 1 } {  \sqrt{\alpha_t} }
\left(
\bx_{t}
-
 \frac{ 1 - \alpha_t} {   \sqrt{1-\oa_t} } \bar{\beps}_t 
 \right).
\label{eq:muq3}
    \end{equation}

The missing ingredient here is $\bar{\beps}_t $---the noise that drove the transition from $\bx_0$ to $\bx_t$.
To deal with this we will train a neural network to predict $\bar{\beps}_t $.
After training, the network will be a black box which takes as input  
\begin{itemize}
   \item  a value of $t$ and a noisy image $\bx_t$
\end{itemize}
 and returns
 \begin{itemize}
   \item  a prediction of $\bar{\beps}_t $.
\end{itemize}
We will denote the prediction by the function $\bepstheta$, where $\theta$ represents the parameters in the neural network---these will be 
learned during the training phase. In each training step, we select an image $\bx_0$ from the training set, take a Gaussian  
$\bar{\beps}_t $ and form a sample of $\bx_t$ using (\ref{eq:f2}). The job of the network is to make the output $\bepstheta$ as close as possible to 
$\bar{\beps}_t $. 

Recalling the expression  (\ref{eq:Nmq}) for the required transition density, 
using the neural network prediction $\bepstheta$ in the expression (\ref{eq:muq3}) for the mean, and adjusting the variance using (\ref{eq:sigmasq}), we will 
obtain $\bx_{t-1}$ from 
\begin{equation}
\bx_{t-1} = 
\frac{ 1 } {  \sqrt{\alpha_t} }
\left(
\bx_{t}
-
 \frac{ 1 - \alpha_t} {   \sqrt{1-\oa_t} }  \, \bepstheta
 \right)
 + \sigma_q(t) \, \bz,
\label{eq:transback}
    \end{equation}
    where $\bz$ is a standard Gaussian.
    This allows us to run the denoising process from $t=T$ to $t=0$.

Having set up the required expressions, in the next section we outline the resulting training and sampling algorithms.

\section{Algorithms}\label{sec:algs}

The training process is summarized in Algorithm~\ref{alg:forward}.
Here we are applying a basic stochastic gradient method \cite{HHdl2019}; in step 5 the network parameters are updated using a least-squares loss function applied to a single, randomly chosen training image.
This simple least-squares formulation can be justified from a likelihood perspective \cite{Cao23,Ho20,Ho22,Sahar22}.
The network architecture used for the experiments in section~\ref{sec:illust} combines 
residual and 
attention blocks
in a U-Net \cite{Ron15} type structure, motivated by the choice in \cite{Ho20}. 
Overall, that network has 12.9 Million parameters across 205 layers.

\begin{algorithm}
\caption{Training with the forward process \cite{Ho20}}\label{alg:forward}
\begin{algorithmic}[1]
\Repeat
\State $\bx_{0}  \sim q(\bx_{0})$  \Comment{choose an image from training set}
\State $t \sim \mathrm{Uniform}(\{1,2,\ldots,T\})$  
\State $\beps \sim \mathrm{N}(\bzero,\bI)$ \Comment{standard Gaussian sample}
\State Take gradient step w.r.t. $\theta$ on $\| \beps - \beps_{\theta}(  \sqrt{\oa_{t}} \, \bx_{0} + \sqrt{ 1 - \oa_{t}} \, \beps, t )\|_2^2$
\Until converged
\end{algorithmic}
\end{algorithm}

Algorithm~\ref{alg:backward} summarizes the sampling process.
Here we define $\sigma_q(1) = 0$, so that only the mean estimate based on (\ref{eq:muq3}) is used at $t=1$.

\begin{algorithm}
\caption{Sampling with the backward process \cite{Ho20}}\label{alg:backward}
\begin{algorithmic}[1]
\State $\bx_{T} \sim \mathrm{N}(\bzero,\bI)$ \Comment{standard Gaussian sample} 
\For{$t = T, T-1, \ldots, 1$}
\State $\bz  \sim \mathrm{N}(\bzero,\bI)$ \Comment{standard Gaussian sample} 
\State $\bx_{t-1} = 
\frac{ 1 } {  \sqrt{\alpha_t} }
\left(
\bx_{t}
-
 \frac{ 1 - \alpha_t} {   \sqrt{1-\oa_t} }  \, \bepstheta
 \right)
 + \sigma_q(t) \, \bz$
\EndFor
\State \Return $\bx_{0}$
\end{algorithmic}
\end{algorithm}



\section{Furthermore}\label{sec:further}

In this section we touch upon some issues that may have occurred to the reader, and provide references where further information may be found.

\noindent{\textbf{How do we judge the performance of generative AI?}}
A generative model must balance the contradictory aims of producing outputs that are 
plausible (like the training data) and novel (not like the training data).
Any attempt to quantify performance must involve somewhat arbitrary choices that allow this circle to be squared.
A popular quantitative measure, which focuses on the plausibility aspect, is 
Fr{\'e}chet Inception Distance \cite{Heu17}.
This measure approximates and compares the probability distributions of the real and synthetic image spaces, under a Gaussian assumption.
Some studies also make use of subjective human opinions, which raises new issues, including reproducibility and representativeness.

\smallskip

\noindent{\textbf{What are useful applications of diffusion models?}}
Given that the internet already stores a bewildering array of real images, it is 
reasonable to ask whether the world needs synthetic examples, however realistic.
However, in some domains representative artificial data is valuable. 
In medical imaging,
for example,  synthetically generated data may help address scarcity, class imbalance and privacy concerns
in educational settings
\cite{Kaz23}.
Perhaps the biggest attraction of diffusion models lies in their use within larger systems.
A diffusion model for image generation may be viewed as a representation of the hidden, or latent, 
distribution of real-world images. 
By conditioning or guiding the image generation according to user-specified requirements, it is then possible to tailor the output to meet certain goals 
\cite{Ba23,goza23,Ho21,Zhang23}.
For example, diffusion forms part of several systems with 
text-to-image capabilities, including 
Open AI's  
DALL-E 2 \cite{ramesh2022hierarchical}, Stability.ai's Dreamstudio \cite{rom22} 
and Google's Imagen \cite{Sahar22}. 
In-painting and overwriting unwanted pixels is also possible \cite{Bur23,Po23}.

Stable diffusion may also be exploited within ChatGPT-style large language models; an example is
Stability.ai's  
StableLM-3B-4E1T \cite{Tow23}. 

\smallskip

\noindent{\textbf{How computationally expensive is it to train and employ a diffusion model?}}
For the simple low-resolution examples in section~\ref{sec:illust}, using a pretrained network to produce new images
is feasible on a standard desktop machine.
However, high resolution image generation with a state-of-the-art
pretrained diffusion model is  
a ``high resource intensive and slow task that prohibits interactive experience for the users and results in huge computational cost on expensive GPUs''
\cite{Ag23}.
The size of many diffusion based models also raises storage issues: ``generating high-resolution images with diffusion models is often infeasible on
consumer-grade GPUs due to the the excessive memory requirements'' \cite{Po23}. 

Training is greater challenge.
For the examples in section~\ref{sec:illust} we trained the network for 500 epochs in 
under 35 minutes on a single 
NVIDIA GeForce RTX 3090 GPU.
It is reported in 
\cite{Wan23}
that 
training the model in \cite{Dhar21} consumes 150-1000 days of NVIDIA V100 GPU time. 
StableLM-3B-4E1T \cite{Tow23} is a 3 Billion parameter language model trained on 1 Trillion tokens of diverse English and code datasets; a 7 Billion parameter version was later released.
Developing smaller-scale versions of such models, or applying the models to compressed latent spaces, is therefore an active area
\cite{rom22,Yang23}.

In terms of power usage when a trained model is deployed, Luccioni et al.\ 
\cite{Lu23}
estimated that 
``the most carbon-intensive image generation model (stable-diffusion-xl-base-1.0) generates 1,594
grams of CO2 for 1,000 inferences, which is roughly the equivalent to 4.1 miles driven by an average gasoline-powered passenger vehicle.''

\smallskip

\noindent{\textbf{Is it a coincidence that 
(\ref{eq:bxalpha}) and 
(\ref{eq:transback}) 
look similar to a numerical discretization of a stochastic differential equation?}}
It is natural to compare 
(\ref{eq:bxalpha}) and 
(\ref{eq:transback}) with the Euler--Maruyama method 
\cite{HK21}, and indeed there are variations of the forward diffusion model that 
have a direct correspondence with stochastic differential equations
\cite{Ho20,Luo22,Po23,Song21}. 
The reverse process may also be associated with backward stochastic differential equations \cite{Wan23}.

\smallskip

\noindent{\textbf{What about the dark side: ethics, privacy, bias and related concerns?}}
Carlini et al.\ 
\cite{car23} 
showed that diffusion models have a tendency to 
memorize and reproduce training images.
For tests on 
Stable AI 
\cite{rom22}
and
Imagen
\cite{Sahar22}
they were able to
``extract over a hundred near-identical
replicas of training images that range from personally
identifiable photos to trademarked logos.''
Somepalli et al.\
\cite{som23} also found examples where a diffusion model ``blatantly copies'' from training data.
The resulting harms to professional artists are considered in 
\cite{Ji23}; these include ``reputational damage, economic loss, plagiarism and copyright infringement.''
When we move into the realm of text-to-image algorithms there are many further issues to consider, including fairness, toxicity and trust \cite{Hao23}.

The figures in section~\ref{sec:illust} indicate that the output from a simple diffusion model is 
difficult to predict and hence to interpret. 
In particular, very different results can be generated
from the same input. Explainable AI is a serious challenge in this setting.

On a more general note, any machine learning algorithm is likely to reflect the biases, omissions and errors 
in the training set \cite{paul21}.
See
\cite{Hutch21} for a proposed  
 framework for data transparency.

We also mention that discussions around ethics in this field often assume that AI is, or will become, all-powerful, thereby 
overlooking empirical observations that 
these systems may fail to operate as intended---the so-called \emph{fallacy of AI functionality}
\cite{Raj22}.
So, as well as the important question of what tasks \emph{should} AI be used for, we must also ask what tasks \emph{can}
AI reliably perform.
This latter issue is ripe for mathematical and statistical contributions. 

Using generative AI to create content (text, images, music, videos, and so on)
that is difficult or impossible to discriminate from human generated content may allow fakery and conspiracy theories to undermine societal safety and benefits. This begets novel risks that are already upon us, identified  in part by the 
inaugural AI Safety Summit which met at Bletchley Park in November 2023.\footnote{
\small{\url{https://www.gov.uk/government/publications/ai-safety-summit-2023-chairs-statement-2-november/chairs-summary-of-the-ai-safety-summit-2023-bletchley-park}}}
Arguably, some of the decadal data science focus on ethics and privacy should have been redirected towards the societal risk of fake truths and the widespread inability to discriminate between content; and the introduction of bias.  These risks now require an in-depth consideration, as we seek to uncover 
and tackle the full range of possibilities. An understanding of the  mathematical foundations of generative AI methods will be a key to ensuring transparency.

\section{PDEs}
\label{sec:pde}

For many applied mathematicians, diffusion is synonymous with certain parabolic PDEs.
Here we present some speculative material that aims to draw a PDE connection with the process described in section~\ref{sec:forwards}. The notion of continuously re-normalizing a diffusion process takes us outside the realm of standard textbook analysis, and opens up some issues that may be of independent interest. Depending on our choice of basic PDE there are several ways to ensure that the  norm of some derivative of the solution remains unchanged over time. Here we illustrate this general idea by continuously re-scaling to preserve the norm of the gradient of the solution, 
that is, the total variation, over the domain.

\label{subsec:sol}
Consider a real valued field $u(x,t)$, where $x\in \Omega$, a bounded domain in $\RR^d$ with a piecewise smooth boundary, $\partial \Omega$, and time $t\ge 0$, satisfying
\begin{equation}
u_t=\Delta u+r(t)u, \quad x\in \Omega, \quad \nabla u.{\bf n}=0, \quad  x\in\partial \Omega,
\label{eqn11}
\end{equation}
for a suitable given initial condition $u(x,0)=u_0(x)$.
Here $r(t)>0$ is a   {\it shadow} time-dependent variable (akin to a Lagrange multiplier),  which continuously rescales $u$ so that   the $L^2$ norm of the gradient of $u$ is preserved. 
More explicitly,  $r(t)$ must be such that 
\begin{equation}
\int_\Omega  ||\nabla u||^2 \,dx  :{=} \int_\Omega \nabla u.\nabla u \, dx \equiv R \quad   \mathrm{(constant)},\ \  t\ge0.
\label{eqn12}
\end{equation}
Here $R>0$ is determined from the initial condition.    

Taking the gradient of  (\ref{eqn11}), and then forming the scalar product with  $\nabla u$, and integrating over $\Omega$, we obtain
\[
\frac{1}{2} 
\,
\frac{d\ }{d t}
\int_\Omega \nabla u.\nabla u \, dx =  
\int_\Omega  {\nabla u. \nabla (\Delta u)}\, dx + 
r(t) \int_\Omega {\nabla u .\nabla u}\, dx.
\]
But, by direct calculation, 
\[
\nabla.(\nabla u \Delta u)=(\Delta u)^2 +\nabla u. \nabla (\Delta u).
\]
So, using the divergence theorem \cite{Matt98}
together with the no-flux boundary condition,  in order to ensure that $R$ is constant we must set 
\[
R \, r(t) =\int_\Omega{(\Delta u)^2}\, dx \ge 0.
\]
Thus we may write (\ref{eqn11}) and (\ref{eqn12}) as the nonlinear integro-differential equation
\begin{equation}
u_t= \Delta u + \frac{u}{R} \int_\Omega{(\Delta u)^2}\, dx, \quad x\in \Omega, \quad \nabla u.{\bf n}=0, \quad  x\in\partial \Omega.
\label{eqn13}
\end{equation}
Here, as before,  $R$ in (\ref{eqn13}) is  set  by the initial condition: $R=\int_\Omega ||\nabla u_0(x)||^2\, dx.$

For any $R>0$,  the constrained equation, (\ref{eqn11}) and (\ref{eqn12}), has infinitely many possible steady states, each of which is of the form
\[
u=\mu_k\phi_k(x) \quad {\rm and} \quad  r=\lambda_k,
\]
where $(\phi_k(x), \lambda_k)$ is the $k$th ($k=0,1,2,...$) eigenfunction-eigenvalue pair for the Laplacian on $\Omega$ with no-flux boundary conditions. However,  $\mu_k$ must satisfy
$$ \mu_k^2= \frac{R}{\int_\Omega  ||\nabla \phi_k||^2 \,dx }, $$
and so  $k\ge 1$, since the simplest eigenfunction satisfies $||\nabla \phi_0||\equiv 0$. 

Now consider small perturbations around the $k$th steady state. We set 
\begin{eqnarray*}
u  &= & \mu_k\left(\phi_k(x) +\epsilon e^{\sigma t} v(x)\right)+O(\epsilon^2),\\
r(t) &=& \lambda_k + \epsilon \beta e^{\sigma t} + O(\epsilon^2),
\end{eqnarray*}
for some  $v(x)$  and constants $(\sigma, \beta)$ to be determined. Substituting these expressions into (\ref{eqn11}), to $ O(\epsilon)$ we obtain  
\begin{equation}
0=\Delta v +(\lambda_k-\sigma)v +\beta \phi_k, \quad x\in \Omega, \quad \nabla v.{\bf n}=0, \quad  x\in\partial \Omega.
\label{eqn4}
\end{equation}

Now setting $v=\phi_{\tilde{k}}$, for $k\ne \tilde{k}$ (since the  eigenfunctions form an orthonormal basis for the solution space), (\ref{eqn4})  yields
\[
0=(\lambda_k - \lambda_{\tilde{k}} -\sigma) \phi_{\tilde{k}} +\beta \phi_k, \quad  x\in \Omega,
\]
so that 
\[
\sigma=\lambda_k - \lambda_{\tilde{k}}  \ {\rm and} \ \beta=0.
\]

The condition on $\sigma $ implies that each steady state is stable with respect to perturbations in all higher eigenmodes, where $k < \tilde{k}$,  yet is unstable with respect to any perturbations in lower eigenmodes, $k > \tilde{k}$.  
Thus over a long time the solution must  decay to the the first eigenmode, $\mu_1 \phi_1(x)$.

\begin{figure}
    \centering
    \includegraphics[width=0.7\textwidth]{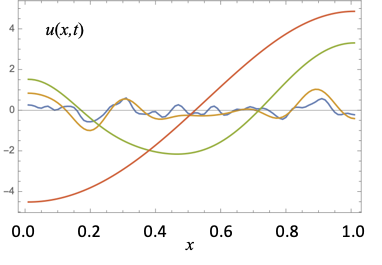}
    \caption{Numerical solution profiles of (\ref{eqn11}) and (\ref{eqn12}) where $d=1$ and $\Omega=[0,1]$, shown at successive times   on a $\log$ timescale ($t=$ 1.5 (blue), 15 (yellow), 150 (green), and 1500 (red)), where any   $\phi_0$ component  has been  set to zero.  In this case each profile has $\int_\Omega  ||\nabla u||^2 \,dx   \equiv R \approx 11.15$. The $\phi_1$ mode dominates at larger times. 
    }
    \label{plotter}
\end{figure}

This is borne out in the numerical experiment shown in Figure \ref{plotter}. Here, for a one dimensional domain $\Omega=[0,1]$,  we have $\phi_k(x)= \sqrt{2}\cos k\pi x$ and $ \lambda_k= k^2\pi^2$.

This framework and others in physics (see for example \cite{xu2022poisson, xu2023pfgm}) suggests a number of ways that generative AI might exploit, or be 
explained by, PDE theory. The central challenge is being able to run the processes in reverse in order to generate plausible content from randomness.

Of course in the example above the parabolicity of the forward dynamic evolution, (\ref{eqn11}) and (\ref{eqn12}), means that, formally, the backward equation is ill-posed, there being no global solution guaranteed, with both discontinuities and point masses possibly occurring as time moves backwards. Hence any backward approximation should require some mollification, perhaps via a numerical solution that leverages finite resolution and norm preserving properties.

In general one should also wish to consider (i)  how these norm-preserving (re-scaling) diffusion systems work when they are subject to stochastic forcing over a finite time, and (ii) how such processes might  be run backwards in seeking to recover or approximate the initial conditions under various Markovian assumptions.

\section*{Data Statement}
Code for the experiments presented here will be made available upon publication.

\bibliographystyle{siam}
\bibliography{diff_refs}

\end{document}